\documentclass{sig-alternate}

\usepackage{hyperref}
\usepackage{graphicx}
\usepackage{amsmath}
\usepackage{amsfonts}

\usepackage{floatrow}
\makeatletter
\def\@copyrightspace{\relax}
\makeatother

\toappear{\the\boilerplate\par
{\confname{\the\conf}} \the\confinfo\par \the\copyrightetc}

\newfont{\mycrnotice}{ptmr8t at 7pt}
\newfont{\myconfname}{ptmri8t at 7pt}
\let\crnotice\mycrnotice%
\let\confname\myconfname%

\begin{document}

\title{A Linked Data Scalability Challenge: \\ Concept Reuse Leads to Semantic Decay}

\numberofauthors{3}

\author{
\alignauthor
Paolo Pareti\\
\affaddr{University of Edinburgh\\United Kingdom}\\
p.pareti@sms.ed.ac.uk
\alignauthor
Ewan Klein\\
\affaddr{University of Edinburgh\\United Kingdom}\\
ewan@inf.ed.ac.uk
\alignauthor
Adam Barker\\
\affaddr{University of St Andrews\\United Kingdom}\\
\mbox{adam.barker@st-andrews.ac.uk}
}

\maketitle

\begin{abstract}
The increasing amount of available Linked Data resources is laying the foundations for more advanced Semantic Web applications. One of their main limitations, however, remains the general low level of data quality. In this paper we focus on a measure of quality which is negatively affected by the increase of the available resources. We propose a measure of semantic richness of Linked Data concepts and we demonstrate our hypothesis that the more a concept is reused, the less semantically rich it becomes. This is a significant scalability issue, as one of the core aspects of Linked Data is the propagation of semantic information on the Web by reusing common terms. We prove our hypothesis with respect to our measure of semantic richness and we validate our model empirically. Finally, we suggest possible future directions to address this scalability problem.
\end{abstract}



\section{Introduction}
\noindent Recent years have seen a constant increase in the amount of available Linked Data resources. While the problem of data availability is gradually reducing, data quality remains one of the main limiting factors of Linked Data applications. At the root of this problem lies the open nature of the Web, where knowledge can be erroneous, incomplete, and where concepts can be used in different ways between different sources. 
In this paper we focus on a particular measure of quality, namely the semantic richness of Linked Data concepts. Following a number-of-features approach \cite{Pexman2008SemanticRichnessNOF}, we define our measure of semantic richness as the amount of facts that we can expect to infer from a concept.
\par Web-scale reuse of semantic formalisations, such as concepts and relations, is an essential aspect of Linked Data \cite{Bizer2009LinkedDataSoFar}. However the openness of Linked Data results in the paradox that, while the reuse of terms is considered a good practice, it also progressively decreases the semantic richness of the reused terms. This trend of diminishing semantics as the usage of a term increases is a severe limitation to the scalability of Linked Data. In fact, the less semantically rich a concept becomes the less assumptions can be safely made about its entities, thus reducing its potential applications.
\par For example, the concept of \emph{Person} in a restricted domain, such as an individual organisation, might follow some particular conventions, such as having an office number and an address. However, if we extend the usage of this concept to more and more entities outside the organisation, then it might not be possible to assume that a \emph{Person} has an office number, or maybe even an address. In the Semantic Web, by virtue of the fact that \emph{anybody can say anything about anything}, an entity of type \emph{Person} could be anything. This problem affects any Linked Data term, including relations. A notable example is the \emph{owl:sameAs}\footnote{\url{http://www.w3.org/2002/07/owl\#sameAs}} relation, originally intended to represent strict equality between two entities, which is frequently misused to represent a wide range of weaker similarity relations \cite{Halpin2010owlSameAsNotTheSame}.
\par The main contributions of this work are two. The first is the validation of our hypothesis that the more datasets reuse a Linked Data concept, the less semantically rich it becomes. The second is a measure of semantic richness that can be used to compare the usage of Linked Data concepts between different datasets. 
This problem will be discussed in more detail in Section \ref{problem_definition} and it will be contextualised with previous research in Section \ref{background}. In Section \ref{semantic_measure} we define our measure of semantic richness and we prove our hypothesis with respect to this measure. In Section \ref{experiments} we validate our hypothesis empirically to provide insights on the extent of this problem. We will then suggest possible future directions to address this problem in Section \ref{methodology}.

\section{Problem Definition}
\label{problem_definition}
\noindent We define a measure of semantic richness $\Gamma$ as a function that takes a concept $\alpha$, such as \emph{foaf:Person}\footnote{\url{http://xmlns.com/foaf/0.1/Person}} and a Linked Data graph $S$, such as the DBpedia graph \cite{Auer2007DBpedia}, and returns a positive real number. This number quantifies the amount of information conveyed by the concept $\alpha$ within the graph $S$. In other words, it is a measure of how many facts we can expect to infer about the instances of that concept that are found in the graph. If $\Gamma(\alpha,S_{1}) > \Gamma(\alpha,S_{2})$ then knowing that an entity is of type $\alpha$ entails more facts in the context of the graph $S_{1}$ rather than $S_{2}$.
\par When creating such measure it is important to consider scalability and robustness, as there is large variability in the size and quality of Linked Data resources. It should be possible to compute this measure effectively over large amounts of Linked Data. Moreover, this measure should be applicable to any dataset, even when taxonomic information is not available or when all the available entities are of a single type. These requirements prevent the applicability of existing semantic measures such as the Information Content.
\par Being based on the usage of a concept, rather than on its position within a conceptual framework such as a taxonomy, this notion differs from the notion of semantic specificity. For example, while a concept is less semantically specific (or less informative) than its sub-concepts, it might be more semantically rich. This situation occurs when the probability of an entity being a member of a sub-concept depends on the other sub-concepts it is a member of, such as in case of mutual exclusivity of sub-concepts. For example, in a dataset $S$ a semantically rich concept $\alpha$ might contain a small set of entities $s$ of which nothing is known. If we consider those entities as members of concept $\beta$, we will have that 
$\Gamma(\beta,S) < \Gamma(\alpha,S)$ although $\beta$ is a sub-concept of $\alpha$.
\par We define $S^{\alpha_{1}}$ and $S^{\alpha_{2}}$ as the sets of entities of type $\alpha$ contained, respectively, in datasets $S_{1}$ and $S_{2}$. Our hypothesis is that the reuse of a Linked Data concept results in a decrease of its semantic richness. Assuming that the sets of entities of type $\alpha$ in two different datasets are disjoint ($S^{\alpha_{1}} \cap S^{\alpha_{2}} = \emptyset$), we express this hypothesis as follows:
\begin{equation}
\label{eq:hypothesis}
 \Gamma(\alpha,S_{1} \cup S_{2}) \leq \frac{|S^{\alpha_{1}}|\Gamma(\alpha,S_{1})+ |S^{\alpha_{2}}|\Gamma(\alpha,S_{2})}{|S^{\alpha_{1}}|+|S^{\alpha_{2}}|}
\end{equation}
Intuitively, this inequation states that the semantic richness of a concept with respect to the union of two graphs is at most equal to the average semantic richness of the graphs. The difference between these values represents the amount of information about a concept that is lost when merging two datasets. In Section \ref{experiments} we will show how, in practice, the union of two graphs typically results in a significant decrease of semantic richness.

\section{Background}
\label{background}
\noindent The notion of Information Content (IC) has been proposed to measure the informativeness of concepts within a taxonomy in the field of Information Theory \cite{Resnik1995InformationContent}. This measure has been shown to be effective not only with respect to concepts, but also to measure the informativeness of resources and relations within a knowledge base \cite{Meymandpour2013LinkedDataInformativeness}. The IC of a concept $\alpha$ within a knowledge base is defined as $IC(\alpha) = -log(p(\alpha))$, where $p(\alpha)$ is the probability of an entity belonging to concept $\alpha$. One limitation to the applicability of this measure is that the probability $p(\alpha)$ might not be meaningful when considering domain specific datasets. For example, several existing datasets contain only instances of type \emph{foaf:Person}. The IC of this concept computed over these datasets will always be equal to $0$ since $p(\text{\emph{foaf:Person}})=1$. The IC measure would not distinguish between the different datasets although they might have a different semantic interpretation of the concept \emph{foaf:Person}. A different measure is therefore necessary to compare the amount of information conveyed by the concept in each dataset.
\par In the field of Cognitive Neuroscience several measures of semantic richness have been proposed \cite{Pexman2008SemanticRichnessNOF}. These measures have been used to explain why certain concepts can be processed better or faster by humans to perform certain tasks.
We adopt one such measure, called number-of-features (NF), by measuring the number of facts that characterise a Linked Data concept. The basic intuition behind the NF measure is that semantically rich concepts tend to be characterised with more attributes than less semantically rich ones. 
\par Our metric to compute the semantic richness of a concept does not assume the availability of pre-existing inference rules. In order to calculate how many facts we can infer about the entities of a given concept we adopt an Inductive Learning approach. For example, if all the entities of concept $\alpha$ from a graph $S$ share property $i$, then we can induce the rule that if an entity belongs to $\alpha$ then it will have property $i$. The applications of Inductive Learning for Linked Data have received considerable attention in the recent years thanks to an increasing availability of Linked Data resources. The main advantages of Inductive Learning approaches are scalability and robustness to errors and uncertainty \cite{Amato2010InductiveLearning}.

\par Schema information can be inferred from a knowledge base by considering the frequency of certain patterns in the data, such as the common properties shared by the entities of the same type \cite{Lorenz2013KnowledgeEnrichment}. This schema information can then be used to evaluate the quality of the knowledge base. For example, it is possible to evaluate how well a Linked Data graph matches a set of schema rules by using SPARQL query patterns \cite{Kontokostas2014LinkedDataQalityTests}. 

\par When considering data from multiple sources, the decrease of semantic richness of a concept can be alleviated by annotating data with contextual information. Several approaches have been proposed to represent contextual information about Linked Data, in particular with respect to Provenance \cite{Moreau2010WebProvenance}. For example, we might be able to infer more knowledge about a set of entities of a concept if we know that they all originate from DBpedia. One limitation of this approach is the difficulty of creating and maintaining contextual information. Moreover, contextual information does not explicitly describe how a concept is used within a dataset.

\section{The Semantic Richness Measure}
\label{semantic_measure}
\noindent Given a concept $\alpha$, such as \texttt{foaf:Person}, we define a measure of its semantic richness $G$ which quantifies the amount of information conveyed by the concept. Unlike IC-based measures \cite{harispe2013semanticMeasures}, our measure is not proportional to the probability of an entity being a member of the concept. This probability cannot be estimated reliably for Linked Data due to its open nature and partial observability. When evaluating the semantic richness of a concept we only assume the availability of a representative subset of entities of that concept obtained from a finite number of sources. Taxonomical information about the concept is not required.
\par Our measure of semantic richness $G$ is defined as a function of the number of facts that we can infer about its instances, and about their probability of being correct. These facts are represented as graph patterns. For example, the fact that persons have a birth date can be represented by the graph pattern: \texttt{\{ ?person a foaf:Person .\ ?person foaf:birthday ?date .\ \}} here represented as a SPARQL\footnote{\url{http://www.w3.org/TR/rdf-sparql-query/}} graph pattern using the FOAF\footnote{\url{http://xmlns.com/foaf/spec/}} ontology. We define the probability of an entity of type $\alpha$ from dataset $S$ matching pattern $i$ as $p(i,\alpha,S)$. The average number of patterns that we can observe in the entities of type $\alpha$ is the expected value $\mu$ of a Poisson binomial distribution of the probabilities $p$. We only need to compute this value over the finite set of patterns $I$ that have been observed among the entities of the dataset, namely when $p(i,\alpha,S) > 0$. The expected value of this distribution can then be computed as follows:
\begin{equation}
\label{expected_value_of_facts}
\mu = \sum_{i \in I}p(i,\alpha,S)
\end{equation}
Equation \ref{expected_value_of_facts} gives us a measure of how many patterns, in average, we can observe from the entities of type $\alpha$ in dataset $S$. This measure, however, does not tell us how well we can characterize a concept with a set of common features. In fact, a high value of $\mu$ could be equally a result of a large number of infrequent patterns or of a small number of frequent patterns. However, frequent patterns are more useful than infrequent ones for the purpose of characterising a concept, for example to learn schema information using Inductive Learning. Taking this into consideration, we define our measure of semantic richness with respect to a set of characteristic patterns $Y \subseteq I$ that define which patterns we can expect entities of type $\alpha$ to have. 
\par If a pattern $i$ belongs to $Y$, then whenever we observe an entity $\epsilon$ of type $\alpha$ we can infer that it would match pattern $i$. We base our measure of semantic richness on the correctness of such inferences. If an entity $\epsilon$ really matches pattern $i$ then we count it as a correct inference, we count it as incorrect otherwise. For example, we might consider including the pattern ``has a dedicated Wikipedia page" in the set of characteristic patterns of concept \emph{foaf:Person}. Whenever we observe an entity of type \emph{foaf:Person} which has a dedicated Wikipedia page we say that inferring this pattern is correct; we say it is incorrect otherwise. 
\par Our measure of semantic richness $G(\alpha,S)$ of a concept $\alpha$ with respect to a dataset $S$ and a set of characteristic patterns $Y$ is the difference between the expected number of correct inferences and the expected number of incorrect ones that we can make about an entity:
\begin{equation}
\label{goal_pre}
	\begin{array}{l}
		G(\alpha,S) = \sum_{i \in Y}p(i,\alpha,S) - \sum_{i \in Y}(1-p(i,\alpha,S)) \\[2mm]
		G(\alpha,S) = \sum_{i \in Y}2p(i,\alpha,S) - 1
	\end{array}
\end{equation}

\noindent From equation \ref{goal_pre} it follows that, in order to maximize the value of $G(\alpha,S)$, $Y$ should be the set of patterns which occur in the majority of the entities ($\forall i \in Y. p(i,\alpha,S) > 0.5$).  Therefore equation \ref{goal} can be rewritten as follows:
\begin{equation}
\label{goal}
	\begin{array}{l}
		G(\alpha,S) = \sum_{i \in I}
		\begin{cases}
			2p(i,\alpha,S) - 1 & \text{if } p(i,\alpha,S) > 0.5 \\
			0 & \text{otherwise }
		\end{cases}
	\end{array}
\end{equation}

\noindent The particular threshold of $0.5$ is a result of the fact that in this measure, which aims at being generic, correct and wrong inferences carry the same (absolute) weight. Similarly, all patterns carry equal weight. This formulation of semantic richness is not meant to be unique and variations are possible. For example, equation \ref{goal} could be modified to penalize wrong inferences more. Also, different patterns could be given different weights to represent how important or informative they are for a specific application.
\par In this work we use use our generic measure of semantic richness to analyse our hypothesis, namely that as we consider more entities as members of a concept, the semantic richness of that concept decreases. In fact, assuming that the two sets of entities are disjoint ($S^{\alpha_{1}} \cap S^{\alpha_{2}} = \emptyset$) the semantic richness of the union of two sources of data is always inferior or equal to the average individual semantic richness of the sources:
\begin{equation}
\label{diminishing_semantics}
G(\alpha,S_{1} \cup S_{2}) \leq \frac{ |S^{\alpha_{1}}|G(\alpha,S_{1}) + |S^{\alpha_{2}}|G(\alpha,S_{2}) }{|S^{\alpha_{1}}|+|S^{\alpha_{2}}|}
\end{equation}

\subsection{Proof of Semantic Richness Decrease}
\label{proofs}
\noindent Formula \ref{diminishing_semantics} can be proved according to our definition of semantic richness defined in Formula \ref{goal}. It can be observed that the semantic richness of a concept is the sum of the individual semantic richness of each pattern $G^{\prime}(\alpha,S,i) = ( 2p(i,\alpha,S) - 1 ) [p(i,\alpha,S) > 0.5]$. Therefore Formula \ref{diminishing_semantics} can be proved by proving that the inequation holds for every individual pattern:
\begin{equation}
\label{diminishing_semantics_proof_goal}
G^{\prime}(\alpha,S_{1} \cup S_{2},i) \leq \frac{ |S^{\alpha_{1}}|G^{\prime}(\alpha,S_{1},i) + |S^{\alpha_{2}}|G^{\prime}(\alpha,S_{2},i) }{|S^{\alpha_{1}}|+|S^{\alpha_{2}}|}
\end{equation}
Having defined $w_{1} = p(i,\alpha,S_{1})$ and $w_{2} = p(i,\alpha,S_{2})$, then the probability $p(i,\alpha,S_{1} \cup S_{2})$ can be defined as ratio between the number of entities of each set which match pattern $i$ and the number of all the entities in the union of the sets: 
\begin{equation}
\label{diminishing_semantics_proof_goal_2}
p(i,\alpha,S_{1} \cup S_{2}) = \frac{ |S^{\alpha_{1}}|w_{i} + |S^{\alpha_{2}}|w_{2} }{|S^{\alpha_{1}}|+|S^{\alpha_{2}}|}
\end{equation}
Formula \ref{diminishing_semantics_proof_goal} can then be written as follows:
\begin{equation}
\label{diminishing_semantics_proof_goal_3}
	\begin{array}{l}
(2\frac{ |S^{\alpha_{1}}|w_{i} + |S^{\alpha_{2}}|w_{2} }{|S^{\alpha_{1}}|+|S^{\alpha_{2}}|} -1) [\frac{ |S^{\alpha_{1}}|w_{i} + |S^{\alpha_{2}}|w_{2} }{|S^{\alpha_{1}}|+|S^{\alpha_{2}}|} > 0.5] \leq \\[2mm]
\frac{ |S^{\alpha_{1}}|(2w_{1} -1)[w_{1}>0.5] + |S^{\alpha_{2}}|(2w_{2} -1) [w_{2}>0.5]}{|S^{\alpha_{1}}|+|S^{\alpha_{2}}|}
	\end{array}
\end{equation}
From Formula \ref{diminishing_semantics_proof_goal_3} it follows that in case $w_{1} \leq 0.5$ and $w_{1} \leq 0.5$ then $p(i,\alpha,S_{1} \cup S_{2}) < 0.5$ and therefore Formula \ref{diminishing_semantics_proof_goal_3} is reduced to the tautology $0 \leq 0$. This matches the intuition that if a pattern was not frequent in any of the original datasets, it will not be frequent also in the union of the datasets.
\par In case $w_{1} > 0.5$ and $w_{1} > 0.5$ then $p(i,\alpha,S_{1} \cup S_{2}) > 0.5$ and therefore equation \ref{diminishing_semantics_proof_goal_3} becomes:
\begin{equation}
\label{diminishing_semantics_proof_goal_4}
	\begin{array}{l}
2\frac{ |S^{\alpha_{1}}|w_{i} + |S^{\alpha_{2}}|w_{2} }{|S^{\alpha_{1}}|+|S^{\alpha_{2}}|} -1 \leq 
\frac{ |S^{\alpha_{1}}|(2w_{1} -1) + |S^{\alpha_{2}}|(2w_{2} -1) }{|S^{\alpha_{1}}|+|S^{\alpha_{2}}|}
	\end{array}
\end{equation}
This equation can be simplified to the tautology $0 \leq 0$. This matches the intuition that a pattern which was frequent in both of the original datasets will still be frequent in the union of the datasets.
\par We then consider the last case in which a pattern is frequent only in one of the original datasets: when $w_{1} > 0.5$ and $w_{2} \leq 0.5$. This requires us to consider two sub-cases, depending on the value of $p(i,\alpha,S_{1} \cup S_{2})$. If $p(i,\alpha,S_{1} \cup S_{2}) \leq 0.5$, then Formula \ref{diminishing_semantics_proof_goal_3} can be simplified to:
\begin{equation}
\label{diminishing_semantics_proof_goal_5}
	\begin{array}{l}
0 \leq 
\frac{ |S^{\alpha_{1}}|(2w_{1} -1)}{|S^{\alpha_{1}}|+|S^{\alpha_{2}}|}
	\end{array}
\end{equation}
This formula can be further simplified to $w_{1} > 0.5$ which is true by assumption. Finally, in case  $w_{1} > 0.5$ and $w_{2} \leq 0.5$ and $p(i,\alpha,S_{1} \cup S_{2}) > 0.5$ Formula \ref{diminishing_semantics_proof_goal_3} can be simplified to:
\begin{equation}
\label{diminishing_semantics_proof_goal_6}
	\begin{array}{l}
2\frac{ |S^{\alpha_{1}}|w_{i} + |S^{\alpha_{2}}|w_{2} }{|S^{\alpha_{1}}|+|S^{\alpha_{2}}|} -1 \leq 
\frac{ |S^{\alpha_{1}}|(2w_{1} -1) }{|S^{\alpha_{1}}|+|S^{\alpha_{2}}|}
	\end{array}
\end{equation}
This equation can be simplified to $w_{2} \leq 0.5$ which is true by assumption. Having proved Formula \ref{diminishing_semantics_proof_goal}, and therefore Formula \ref{diminishing_semantics}, we have proved our hypothesis (Formula \ref{eq:hypothesis}) with respect to our measure of semantic richness.

\section{Experimental Evaluation}
\label{experiments}
\noindent Having defined our model to measure semantic richness we perform an empirical validation of this measure and of our hypothesis. This experimental evaluation, performed over several large datasets, also provides evidence of the scalability of our measure. These experiments are based on an implementation of a system that can sample entities from a dataset and calculate the semantic richness of a concept within the dataset.

\begin{figure}[tb]
			\centering
			\includegraphics[width=1.00\textwidth]{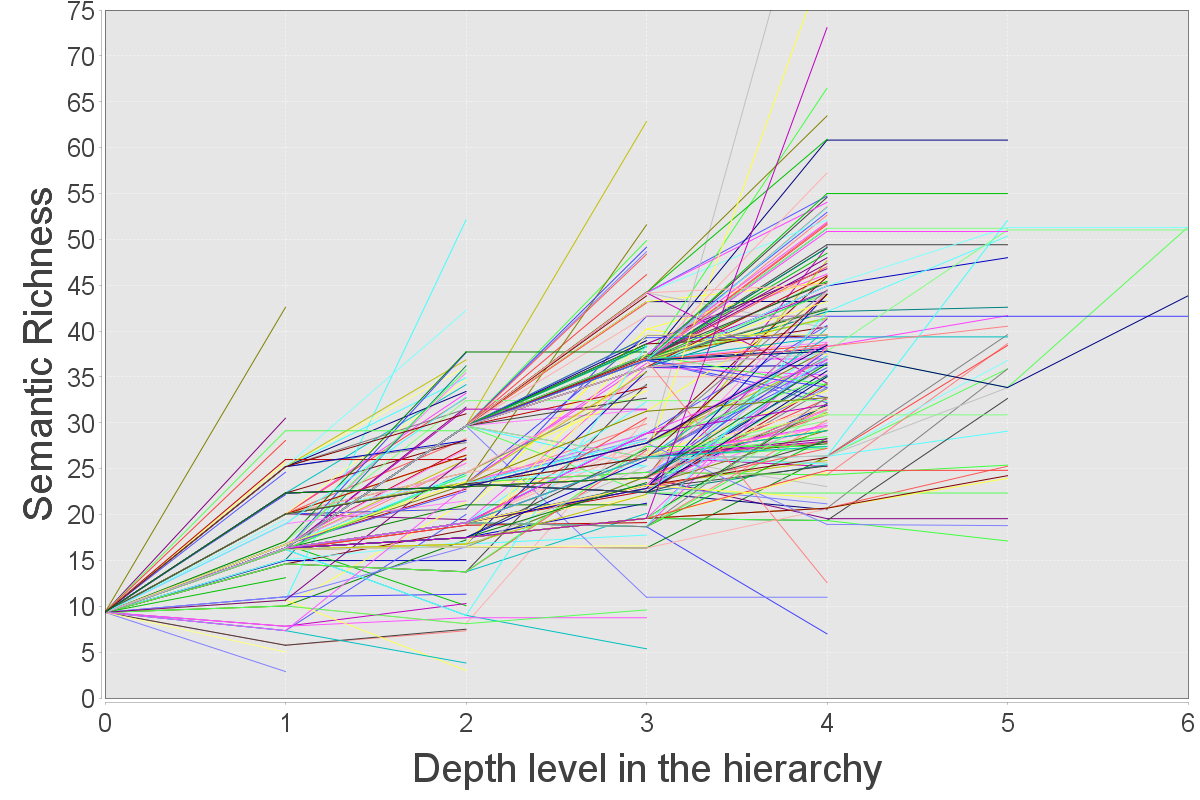}
			\vspace{-4mm}
	\caption{The DBpedia ontology tree represented in function of the semantic richness of each concept. Each segment represents a subclass relation.}
	\label{fig:increase_of_G}
\end{figure} 

\subsection{Validation of the Semantic Measure}
\noindent Our approach uses a numerical measure of the semantic richness of a concept using equation \ref{goal}. But how well does this measure actually capture the concept of semantic richness? We evaluate this measure according to the intuition that concepts are, in average, less semantically rich than their sub-concepts. A similar property is also found in the IC measure, as the IC of a concept is, by definition, always inferior or equal to the IC of its sub-concepts.
\par We have evaluated how well our measure matches this intuition on the concepts of the DBpedia ontology\footnote{\url{http://dbpedia.org/ontology/}} by calculating the semantic richness of the entities of each concept.\footnote{The entities were extracted from DBpedia version 3.5.1} To reduce the computational complexity, all concepts with over 1000 entities have been replaced by a randomly sampled set of 1000 entities. Also, we have considered only simple graph patterns which involve a single triple. More specifically, we have considered patterns as pairs $<p,o>$ which match an entity $\epsilon$ if the triple $<\epsilon, p, o>$ is asserted in the dataset. Improving the quality of this measure by considering more complex patterns is left for further research.
\par Figure \ref{fig:increase_of_G} shows the change of semantic richness between each concept and its sub-concepts. This graph shows a significant tendency of concepts to be less semantically rich than their sub-concepts. In fact, 90\% of the subclass relations resulted in an increase of semantic richness from a concept to its sub-concepts. Also, in all cases concepts resulted in a lower semantic richness than the average semantic richness of their sub-concepts, thus supporting our hypothesis. Only in 10\% of the cases a subclass relation resulted in a decrease of semantic richness from a concept to a sub-concept. This situation occurred because in the DBpedia ontology, sibling concepts are mutually exclusive. Therefore, the knowledge that an entity belongs to a certain concept affects the probability of it being a member of another, potentially more semantically rich, concept.

\subsection{Validation of the Hypothesis}
\noindent We have run an experiment to quantify the decrease in semantic richness as concepts are reused. In this experiment we have considered the concept \emph{foaf:Person} as it is used in ten different datasets. The list of the endpoints used to access the datasets is shown in Table \ref{tab:sources}. From each endpoint we have extracted information about 1000 randomly selected entities.
\par
\begin{figure}[tb]
			\centering
			\includegraphics[width=1.00\textwidth]{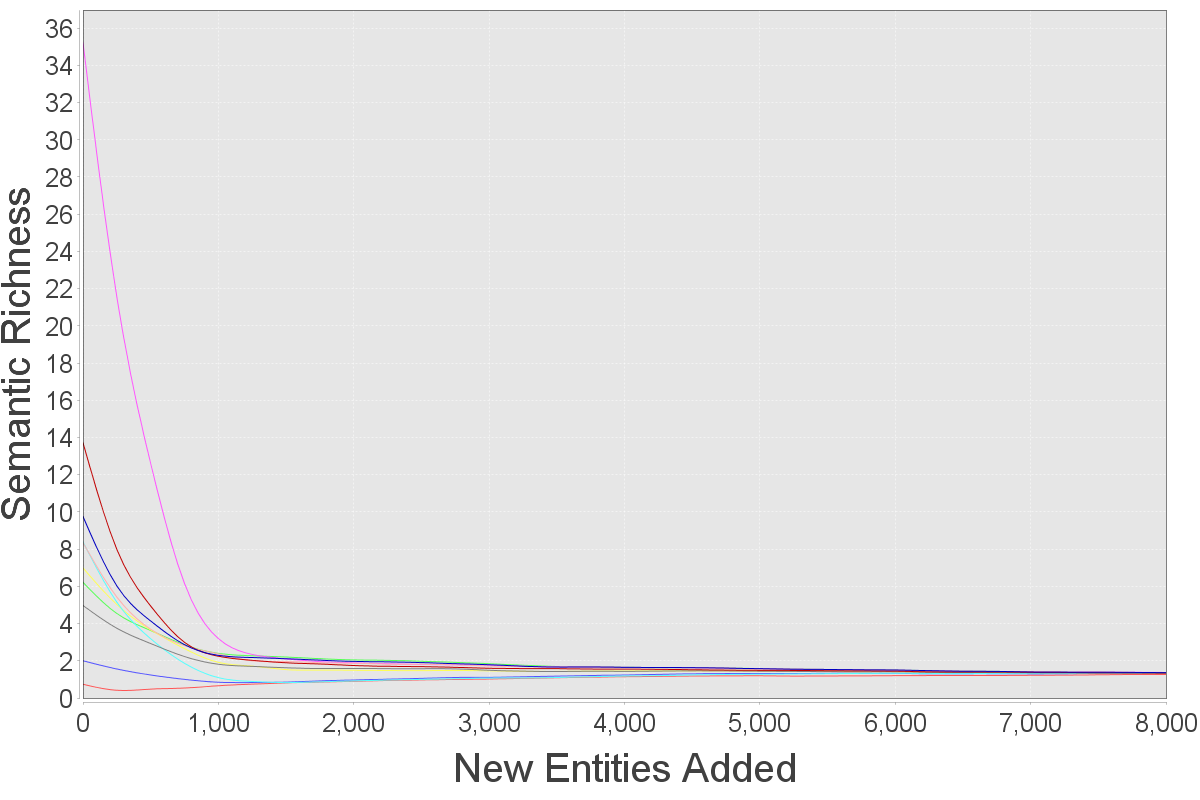}
			\vspace{-4mm}
	\caption{Decrease of semantic richness of the concept \emph{foaf:Person} of ten different datasets as we add more entities from the other datasets.}
	\label{fig:decay_of_G}
\end{figure}

 For each dataset, we have calculated its semantic richness at different states. Starting from the semantic richness of the original dataset, we have recalculated this measure after adding progressively more entities from the other datasets. Figure \ref{fig:decay_of_G} shows the progressive decay of semantic richness as we add new entities of the same type from other sources. The semantic richness of each dataset quickly converges to a value which is significantly inferior to the average measure. While the average semantic richness of all the individual sources is $9.6$, the semantic richness of the merged set results is $1.3$. This experiment validates our hypothesis since the semantic richness of the union of a number of sources is inferior to their average individual semantic richness. Another fact that can be observed is that different datasets represent the same concept with significantly different levels of semantic richness. In this case, the values ranged from $0.7$ of \url{services.data.gov.uk} to $35.2$ of \url{dbpedia.org}.

\begin{table}[tp]
	\centering
		\begin{tabular}{ | l | }
			\hline
			http://dbpedia.org/sparql \\ \hline
			http://data.nobelprize.org/sparql \\ \hline
			http://data.archiveshub.ac.uk/sparql \\ \hline
			http://dati.camera.it/sparql \\ \hline
			http://data.utpl.edu.ec/ecuadorresearch/lod/sparql \\ \hline
			http://lod.sztaki.hu/sparql \\ \hline
			http://semantic.eea.europa.eu/sparql \\ \hline
			http://data.open.ac.uk/query \\ \hline
			http://semanticweb.cs.vu.nl/europeana/sparql \\ \hline
			http://services.data.gov.uk/reference/sparql \\ \hline
		\end{tabular}
	\caption{SPARQL endpoints used in the experiment (accessed on 7/3/2015)}
	\label{tab:sources}
\end{table}

\section{Retaining Semantic Richness}
\label{methodology}
\noindent Since Linked Data does not allow negation, it is not possible to prevent an entity from becoming a member of a concept. It is however possible to predict the effect that adding an entity to a dataset would have on the semantic richness of a concept.
We define $E$ as the set of patterns that are matched by entity $\epsilon$. The degree of typicality of an entity with respect to a concept $\alpha$ can then be defined as $\delta^{\epsilon} = |E \cap Y |/|Y|$ where $Y$ is the set of frequent patterns of $\alpha$. Given two entities $\epsilon$ and $\epsilon^{\prime}$, we can say that $\epsilon$ is more typical than $\epsilon^{\prime}$ if $\delta^{\epsilon} > \delta^{\epsilon^{\prime}}$.
We define an entity to be \emph{atypical}, meaning that including it into a concept would result in a decrease of its semantic richness, if $\delta < 0.5$. If $\delta \geq 0.5$ we say that the entity is \emph{typical}, meaning that its inclusion in the concept would maintain or increase its semantic richness.
\par This typicality measure is scalable and easily verifiable because it can be computed automatically over a subset of the data. In fact, any agent can calculate whether an entity is a typical or an atypical member of a concept. In a closed domain, this measure could be used to determine which entities should be considered as members of a concept, and which should not. In Linked Data, however, it is not possible, nor desirable, to prevent entities from becoming members of a concept. It is however possible to add more knowledge about the typical entities to preserve their semantic richness. A possible approach to do this is to create a sub-concept which groups together the typical entities.
\par For example, the particular way in which the concept \emph{foaf:Person} is used in DBpedia results in the property that the majority of those entities have an associated Wikipedia page. If we merge these entities with another source about persons which are not related with a Wikipedia page, then this property will no longer hold, thus reducing the semantic richness of the concept. In order to preserve this property, we can create a new sub-concept of \emph{foaf:Person} that captures the particular interpretation of this concept adopted by DBpedia.

\section{Conclusion}
\noindent In this paper we have proposed a measure of semantic richness of Linked Data concepts that quantifies the amount of facts that can be inferred from a concept within the scope a particular dataset. This measure differs from previous semantic measures, such as Information Content, as it is meant to compare the difference in semantic richness of the same concept between different datasets. We have validated this measure with respect to the DBpedia ontology by showing that it captures the expected intuition that sub-concepts are, in average, more semantically rich than their super-concepts. We have used the mathematical formulation of this measure to prove our hypothesis that the more a concept is reused, the less semantically rich it becomes. To verify the extent of this problem we have compared the semantic richness of the concept \emph{foaf:Person} from ten different existing datasets. Our preliminary results show that the semantic richness of a concept quickly decreases when merging entities of multiple datasets together. We have also shown that the same concept can be used in different datasets with significantly different levels of semantic richness. Since it is not desirable to prevent concepts from being reused we propose to preserve their semantic richness by organising their heterogeneous set of entities in a number of sub-concepts. These sub-concepts can be generated and populated automatically by considering the effect on their semantic richness that would result after the inclusion of a new entity.

\bibliographystyle{abbrv}
\bibliography{phdbib} 

\end{document}